\documentclass{article} 
\usepackage{collas2023_conference,times}
\usepackage{easyReview}

\usepackage{amsmath,amsfonts,bm}

\def\eqref#1{equation~\ref{#1}}

\def\1{\bm{1}}

\DeclareMathAlphabet{\mathsfit}{\encodingdefault}{\sfdefault}{m}{sl}
\SetMathAlphabet{\mathsfit}{bold}{\encodingdefault}{\sfdefault}{bx}{n}

\usepackage{hyperref}
\hypersetup{
    colorlinks=true,
    linkcolor=red,
    filecolor=magenta,
    urlcolor=blue,
    citecolor=purple,
    pdftitle={Overleaf Example},
    pdfpagemode=FullScreen,
    }

\usepackage{enumitem}
\usepackage{adjustbox} 
\definecolor{blue}{HTML}{0000FF}
\usepackage{booktabs} 

\usepackage{amsmath}
\usepackage{amsthm}
\usepackage{mathtools}
\usepackage{amssymb}

\title{Comparing the Efficacy of Fine-Tuning and Meta-Learning for Few-Shot Policy Imitation}

\author{Massimiliano Patacchiola\\
University of Cambridge\\
\texttt{mp2008@cam.ac.uk} \\
\And
Mingfei Sun \\
University of Manchester\\
\texttt{mingfei.sun@manchester.ac.uk}\\
\AND
Katja Hofmann  \\
Microsoft Research \\
\texttt{katja.hofmann@microsoft.com~~~~~~~~} \\
\And
Richard E. Turner \\
University of Cambridge\\
\texttt{ret26@cam.ac.uk}\\
}

\collasfinalcopy 

\begin{document}

\maketitle

\begin{abstract}
In this paper we explore few-shot imitation learning for control problems, which involves learning to imitate a target policy by accessing a limited set of offline rollouts. This setting has been relatively under-explored despite its relevance to robotics and control applications. State-of-the-art methods developed to tackle few-shot imitation rely on meta-learning, which is expensive to train as it requires access to a distribution over tasks (rollouts from many target policies and variations of the base environment). Given this limitation we investigate an alternative approach, fine-tuning, a family of methods that pretrain on a single dataset and then fine-tune on unseen domain-specific data. Recent work has shown that fine-tuners outperform meta-learners in few-shot image classification tasks, especially when the data is out-of-domain. Here we evaluate to what extent this is true for control problems, proposing a simple yet effective baseline which relies on two stages: (i) training a base policy online via reinforcement learning (e.g.~Soft Actor-Critic) on a single base environment, (ii) fine-tuning the base policy via behavioral cloning on a few offline rollouts of the target policy. Despite its simplicity this baseline is competitive with meta-learning methods on a variety of conditions and is able to imitate target policies trained on unseen variations of the original environment. Importantly, the proposed approach is practical and easy to implement, as it does not need any complex meta-training protocol. As a further contribution, we release an open source dataset called iMuJoCo (iMitation MuJoCo) consisting of 154 variants of popular OpenAI-Gym MuJoCo environments with associated pretrained target policies and rollouts, which can be used by the community to study few-shot imitation learning and offline reinforcement learning.
\end{abstract}

\section{Introduction}

The ability to learn quickly is critical for autonomous agents as it enables them to rapidly adapt to new environments and tasks. For example, in the field of robotics, agents may encounter changes in the environment or control dynamics, leading to a need for a policy update. These changes can occur in contexts such as manipulation \citep{fu2016one}, locomotion \citep{nagabandi2018learning}, human-robot interaction \citep{khamassi2018robot}, and may be caused by a variety of factors. There are many ways in which changes to the transition dynamics can arise, including faulty sensors, parts of the robot being replaced or upgraded over time, the design of the robot changing, parts of the robot corroding or wearing over time, environmental damage (sand in joints, lubricant running low, etc.). In all these cases, adaptation via imitation could be a possible way to adjust the policy on trajectories obtained via corrective feedback provided by humans or other agents.

Robotics and automation are not the only areas where imitation is important. For example, in multi-player games like DOTA~\citep{berner2019dota}, StarCraft~\citep{vinyals2019grandmaster}, and Minecraft~\citep{johnson2016malmo}, adapting a pretrained policy to the style of the team is crucial, and this can be accomplished through imitation. Additionally, in 1-vs-1 gameplay, a similar approach can be used to fine-tune a pretrained policy on the player's trajectories to create a fair opponent calibrated to the user's expertise \citep{paraschos2023game}.

While imitation can be a promising solution for adapting policies, it does have some limitations that need to be considered. One significant restriction is the potential scarcity of samples provided by a target policy, especially when relying on human-generated data. In some cases, obtaining a large number of diverse and informative trajectories can be challenging. When humans provide the trajectories, the availability of samples may be limited due to factors such as time constraints, human resources, or the complexity of the task.

To address these limitations, researchers have been exploring the use of few-shot learning as a potential solution. Few-shot learning has gained significant attention in recent years, as it offers a promising approach for learning from a small amount of task-specific data.  This setting is especially relevant in applications including personalization \citep{massiceti2021orbit}, object detection \citep{kang2019few}, and molecule design \citep{stanley2021fs}. The majority of research has focused on image classification, where unlabeled query images are classified based on a support set of labeled images from the same classes. Historically, two approaches have been employed to tackle the few-shot image classification problem: fine-tuning and meta-learning. Fine-tuning involves taking a pretrained model, such as a convolutional neural network, and adjusting its parameters using a smaller dataset specific to the target task. Fine-tuning allows the model to learn task-specific features while still retaining the high-level features learned during pretraining, leading to improved accuracy and faster convergence on the target task.
Meta-learning is based on learning-to-learn paradigms \citep{hospedales2020meta}. The underlying idea of meta-learning is to improve the learning process of a model by leveraging previous learning experiences to quickly adapt to new tasks or environments with minimal additional data or training. 
Recent work has showed that fine-tuners are better than meta-learners in low-data regimes \citep{kolesnikov2020big, shysheya2023fit}, reporting superior classification accuracy on large benchmarks such as Meta-Dataset \citep{triantafillou2019meta} and VTAB \citep{zhai2019large}. Specifically, pretraining a large backbone, such as ResNet50, on a generic dataset, like ImageNet, provides a robust starting point for fine-tuning routines that enable the model to adapt to specialized datasets. These results have led the community to question the effectiveness and utility of meta-learning in few-shot image classification.

In this paper, we take these considerations into account and assess to what extent they apply in the setting of few-shot imitation learning for control problems, which involves learning to imitate a target policy by accessing an offline set of its rollouts. Although it has received less attention compared to few-shot image classification, this setting holds high relevance due to the potential for developing techniques that can enhance policies in diverse applications, including robotics and autonomous driving. We focus on scenarios where only state-action pairs from a given trajectory over the target policy can be recorded and stored. This is the most challenging and practical scenario as it allows for learning from human demonstrations.

Meta-learning has been previously used to tackle few-shot imitation \citep{finn2017one}. Using meta-learning in this context implies training a meta-policy over a task distribution that includes numerous target policies after which the model is adapted to imitate a new set of policies. Unlike the few-shot image classification case, fine-tuning baselines are difficult to define for few-shot imitation because there is no equivalent of ImageNet that can be used for model pretraining. We believe that this resulted in the underestimation of the potential of fine-tuning methods for few-shot imitation.

In this work, we demonstrate that it is possible to define a fine-tuning method for few-shot imitation by training a base model via reinforcement learning (e.g.~Soft Actor-Critic) on a single environment and we then fine-tune the base model on offline rollouts, effectively generalizing to target policies trained on unseen variations of the base environment. One key advantage of our proposed baseline is that it eliminates the need for an expensive meta-training stage that relies on a task distribution (distribution over many target policies and variations of the environment). This advantage is crucial in many applications, since generating variations of the base environment (e.g.~using domain randomization) may be difficult or costly in many real-world settings during the training phase. In addition, the use of policies trained via standard RL methods, open the door to the possibility of using open-source models that are publicly available and which can then be fine-tuned on new settings.

The \emph{main contributions} of this paper can be summarized as follows:

\begin{enumerate}
    \item We shed more light on few-shot imitation learning. This problem has been relatively under-explored despite its relevance to robotics and control applications. In particular, we conduct a comprehensive comparison of different methods across various environments and different numbers of shots, offering a broad perspective on each method's strengths and limitations in these settings.
    \item We provide an effective fine-tuning baseline which is based on two stages: (i) pretraining a policy on a single environment via online reinforcement learning and (ii) fine-tuning the policy on offline trajectories from an unseen target policy. We empirically show that this baseline is competitive with meta-learning in the medium and high-shot conditions, achieving similar loss on a large number of tasks.
    \item We release an open-source dataset called iMuJoCo (iMitation MuJoCo) which includes 154 environment variants obtained from the popular OpenAI-Gym MuJoCo suite with associated pretrained target policies and trajectories. This dataset can be used to study imitation-learning and offline RL. The dataset is available at: \url{https://github.com/mpatacchiola/imujoco}
\end{enumerate}

\section{Description of the method}

\subsection{Problem setting}

The objective of our research is to develop a base policy that can quickly imitate a target policy given a few demonstrations. We consider a distribution over environments $p(\mathcal{E})$ where each environment $\mathcal{E}_i \sim p(\mathcal{E})$ is a specific Markov Decision Process (MDP). We make the assumption that the environments in $p(\mathcal{E})$ share a set of task-common parameters, but differ in terms of task-specific parameters. For instance, the environments can have the same transition dynamics but differ with respect to the agent's body structure (e.g.~different body mass, limb length, range of motion, etc.). For each environment we have access to trajectories drawn from a target policy $\pi_{i}$ that has been trained on that environment, where we define a trajectory as a sequence of state-action tuples
\begin{equation}
    \tau \triangleq \{(s_1,a_1), (s_2,a_2), \dots, (s_T,a_T) \}.
\end{equation}

Trajectories are grouped into two disjoint sets: the \emph{support set} $\mathcal{S}_i = \{ \tau_1, \dots, \tau_K \}$ and \emph{query set} $\mathcal{Q}_i = \{ \tau_1, \dots, \tau_L \}$, forming a task $\mathcal{T}_i = \{\mathcal{S}_i, \mathcal{Q}_i\}$. The objective of a learner is to find the parameters $\boldsymbol{\theta}$ of a policy $\hat{\pi}_{\theta}$ that can be effectively adapted to imitate a target policy $\pi^{\ast}$ (trained on the unseen environment $\mathcal{E}^{\ast}$) by conditioning on the support trajectories $\mathcal{S}^{\ast}$. Note that this setting is different from offline RL \citep{levine2020offline} since we assume that the rewards are not available (which is a less restrictive assumption), and that we are dealing with small amount of data (few-shot setting).

\textbf{Meta-Learning} 
Meta-learning methods used in this few-shot imitation learning setting \citep{finn2017one} assume that is possible to access a distribution over tasks $\mathcal{T}_i \sim p(\mathcal{T})$ during the training phase. The parameters $\boldsymbol{\theta}$ of the base policy $\hat{\pi}_{\theta}$ are optimized in two stages, first an inner-loss w.r.t.~the support trajectories $\mathcal{S}_i$ is minimized to obtain a set of task-specific parameters, then an outer-loss w.r.t.~the query trajectories $\mathcal{Q}_i$ is minimized by backpropagating the gradients through the inner-loop and updating the original set of parameters.
At evaluation time, given the support set $\mathcal{S}^{\ast}$ from an unseen target policy $\pi^{\ast}$, the parameter of the imitation policy $\hat{\pi}_{\theta}$ are optimized through the inner-loop routine to obtain the task-specific parameters. Note that the initial assumption made in meta-learning, that we can access samples from a distribution over tasks $\mathcal{T}_i \sim p(\mathcal{T})$, may be unrealistic in many settings, as it implies sampling over a large number of trajectories generated by many different policies. This motivates our effort in finding a fine-tuning baseline that is not limited by such a constraint.

\subsection{Method}

\begin{figure}[t]
\centering
\includegraphics[width=\textwidth]{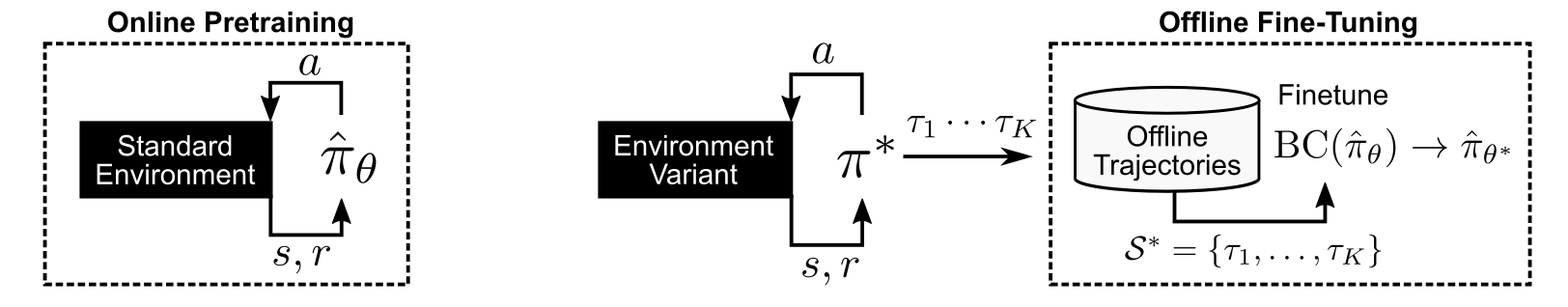}
\caption{Online pretraining and offline fine-tuning of the proposed baseline. During pretraining (left schematic) a base policy $\hat{\pi}_{\theta}$ is trained online using standard RL methods (e.g.~Soft Actor-Critic). In the adaptation phase (right schematic) we have access to an offline support set of trajectories $\mathcal{S}^{\ast}=\{\tau_1, \dots \tau_K \}$ generated by a target policy $\pi^{\ast}$ trained on a variant of the base environment. Offline fine-tuning exploits the offline support-set $\mathcal{S}^{\ast}$ to adjust the parameters of the base policy towards the target via Behavioral Cloning (BC).}
\label{fig_baseline}
\end{figure}

The baseline presented in this paper consists of two phases: 1) online pretraining, and 2) offline fine-tuning. In the first phase, a policy is trained via standard reinforcement learning by accessing a single variation of the environment. In the second phase, the policy trained in the first phase is fine-tuned on the support set of an unseen task. Below we describe these two phases in more detail.

\textbf{Online pretraining} In the pretraining stage we find the parameters of a policy $\hat{\pi}_{\theta}$ using standard Reinforcement Learning (RL).
In standard RL with infinite horizon, the objective to maximize is defined as $J(\pi) = \mathbb{E}_{\pi}\big[ \sum_{t=0}^{\infty}\gamma^t r_t\big]$, 
where $\pi$ is a stochastic policy $\pi: \mathcal{S}\mapsto\Delta(\mathcal{A})$ and  $\gamma\in [0, 1)$ is the discount factor. 
We use Soft Actor-Critic (SAC, \citealt{haarnoja2018soft}) to optimize the policy since it is a state-of-the-art off-policy algorithm for continuous control problems and enjoys good sample efficiency. 
SAC learns a policy $\hat{\pi}_{\theta}$ and a critic $Q_{\psi}$ by maximizing a weighted objective of the reward and the policy entropy term, $\mathbb{E}_{s_t, a_t\sim\hat{\pi}}\big[ \sum_t r_t + \alpha \mathcal{H}\big(\hat{\pi}(\cdot|s_t)\big) \big]$. 
The critic parameters are learned by minimizing the squared Bellman error using transition tuples $(s_t, a_t, s_{t+1}, r_t)$ from a replay buffer $\mathcal{D}$:
\begin{equation}
\mathcal{L}_{Q}(\psi) \triangleq \mathbb{E}_{(s_t, a_t, s_{t+1}, r_t)\sim\mathcal{D}}\big[ Q_{\psi}(s_t, a_t) - r_t - \gamma V(s_{t+1}) \big]^2, 
\end{equation}
where the target value of the next state is obtained by drawing an action $a'$ using the current policy:
\begin{equation}
V(s_{t+1}) \approx \bar{Q}_{\psi}(s_{t+1}, a') - \alpha \hat{\pi}_{\theta}(a|s_{t+1}). 
\end{equation}
The term $\bar{Q}_{\psi}(s_{t+1}, a')$ is usually a copy of the trained $Q_{\psi}$, with parameters that are updated more slowly than $\psi$. 
The policy is learned by minimizing its divergence from the exponential of the soft-$Q$ function, which is equivalent to
\begin{equation}
\mathcal{L}_{\hat{\pi}}(\theta) \triangleq - \mathbb{E}_{a\sim\hat{\pi}}\big[ Q_{\psi}(s_t, a) - \alpha \log\hat{\pi}_{\theta}(a|s_t) \big], 
\end{equation}
where $\alpha$ is the entropy regularization coefficient.

Note that, there is crucial difference between the online pretraining stage used in our method and the meta-training stage used in meta-learning \citep{finn2017model,finn2017one}. Online pretraining only relies on a single base environment to train the imitation policy, whereas meta-learning needs a distribution over many environments and offline trajectories. 

\textbf{Offline Fine-Tuning} After the imitation policy $\hat{\pi}_{\theta}$ has been pretrained on the base environment via SAC, we can use the policy at evaluation time. Given the support trajectories $\mathcal{S}^{\ast} = \{\tau_1, \dots, \tau_K \}$ from an unseen target policy $\pi^{\ast}$, we fine-tune the parameters of the base policy by minimizing a loss function $\mathcal{L}$ over the support points via Behavioral Cloning (BC). This step provides a new set of parameters $\boldsymbol{\theta}^{\ast}$ of the base policy $\hat{\pi}_{\theta^{\ast}}$
\begin{equation}
    \text{BC}(\hat{\pi}_{\theta}, \mathcal{S}^{\ast}, \mathcal{L}) \rightarrow \hat{\pi}_{\theta^{\ast}},
\end{equation}
where we have used the functional notation $\text{BC}(\cdot)$ as a shorthand to represent the iterative gradient-descent routine used in BC. After the offline fine-tuning stage we have access to a specialized policy $\hat{\pi}_{\theta^{\ast}}$ which is an approximation of the target policy $\pi^{\ast}$. Here it is important to stress the fact that offline fine-tuning is part of the evaluation phase, meaning that it has no impact on the training phase. A graphical representation of the online pretraining and offline fine-tuning stages is presented in Figure~\ref{fig_baseline}.

\section{Previous work}

\textbf{Meta-Learning vs.~Fine-Tuning} In the last few years, meta-learning has become a popular approach for tackling the few-shot setting in classification and regression \citep{hospedales2020meta}. One of the most widely used meta-learning methods is Model-Agnostic Meta-Learning (MAML, \citealt{finn2017model}). MAML trains a neural network to find a set of weights that can be easily adapted to new tasks in a few gradient steps. This is achieved through an outer loop that optimizes the neural network weights and an inner loop that updates the weights for a new task. In addition to MAML, other solutions that exploit the learning-to-learn paradigm have been proposed. For instance, metric learning \citep{snell2017prototypical, bronskill2021memory}, Bayesian methods \citep{gordon2018meta, patacchiola2020bayesian, sendera2021non}, and different adaptation mechanisms \citep{rebuffi2017learning, patacchiola2022contextual}. A recent line of work has showed that fine-tuning methods are more effective than meta-learning while being easier to parallelize and train. In the context of few-shot image classification, \cite{chen2019closer} were the first to demonstrate the potential of simple fine-tuning baselines for transfer learning. \cite{triantafillou2019meta} have proposed MD-Transfer as an effective fine-tuning baseline for the MetaDataset benchmark. \cite{kolesnikov2020big} presented Big Transfer (BiT), which shows that large models pretrained on ILSVRC-2012 ImageNet and the full ImageNet-21k are highly effective in transfer learning. Fine-tuning only the last linear layer of a pretrained backbone \citep{bauer2017discriminative, tian2020rethinking} or a subset of the parameters \citep{shysheya2023fit} has also been shown to be a valid approach to transfer learning and few-shot learning.
In the context of reinforcement learning, \cite{mandi2022effectiveness} have compared meta-learning and fine-tuning for adaptation to new environments. This work differs from ours for a number of reasons: (i) it focuses on task-level online adaptation vs.~we focus on offline policy imitation; (ii) it is based on tasks with non-overlapping state-action spaces vs.~we use tasks with overlapping states; (iii) pretraining is performed using a multi-task schedule vs.~pretraining on a single environment condition; (iv) discrete state spaces (vision tasks) vs.~continuous state-action spaces (control tasks).

\textbf{Few-shot imitation learning} \cite{finn2017one} are the first to tackle the imitation problem through a meta-learning approach. The authors adapt MAML to this setting, by assuming that support and query sets (containing state-action tuples) can be sampled from a distribution over many expert policies. \cite{duan2017one} consider a similar setting, assuming access to a very large (possibly infinite) set of tasks, the authors exploit soft attention mechanisms and recurrent architectures to perform adaptation to new tasks. \cite{james2018task} departs from meta-learning and use a metric learning approach for visual imitation in robotics applications.
\cite{yu2018one} aim at performing imitation learning from human demonstrations. The support set consists of states (videos) of a human performing the task, while the query set is represented as a sequence of state-action pairs obtained from a robot that is performing the same task. The challenge here is that the human demonstration only include states (video frames) and there is no easy way to extract the actions (e.g.~joint commands). To solve this problem the authors propose to meta-learn an adaptation objective that does not require actions.
\cite{mitchell2021offline} propose Meta-Actor Critic with Advantage Weighting (MACAW), an optimization-based meta-learning algorithm that uses simple, supervised regression objectives for both the inner and outer loop of meta-training. This work was the first to consider the setting where the state-action-reward tuples arrive as a stream, which is more aligned with the standard offline-RL setting.
\cite{xu2022prompting} exploit prompting in conjunction with a pretrained decision transformer to achieve generalization across continuous control tasks. Few-shot demonstrations are passed as input prompt to a decision transformer, and concatenated with the state-action-reward tuple coming from the interaction with the environment. Like for \cite{mitchell2021offline}, the authors assume an offline-RL streaming containing state-action-reward tuples.
\cite{nagabandi2018learning} consider each timestep to potentially be a new task, and details or settings could have changed at any timestep. The authors present two versions of their method: recurrence-based adaptive learner (ReBAL) and gradient-based adaptive learner (GrBAL). Both methods rely on model predictive control (MPC), which is expensive to run.

\textbf{Generalization in reinforcement learning} Generalization in reinforcement dates back to early works which demonstrated generalization to new tasks in robotics~\citep{sutton2011horde}, and in video games~\citep{schaul2015universal,parisotto2015actor}.
In these applications, the tasks were defined as to reaching a set of goal states~\citep{sutton2011horde}, for which the reinforcement learning was leveraged to minimize the distance to the goal~\citep{andrychowicz2017hindsight,nasiriany2019planning}, with possible combination of language~\citep{jiang2019language}, and the learnt latent spaces~\citep{eysenbach2018diversity,rakelly2019efficient}. 
Multi-task reinforcement learning (MTRL,~\citealt{vithayathil2020survey}) further generalizes the goal reaching to a full set of tasks, which features for example all tasks in Atari~\citep{hafner2019dream}, and various agents with different dynamics or morphology~\citep{huang2020one}. 
Such generalization is challenging because different agents/tasks typically have incompatible observation and action spaces, precluding the direct application of classic deep reinforcement learning approaches. Recent work~\citep{chang2021modularity,goyal2019recurrent} has shown that modularization simplifies learning over different state-action spaces~\citep{huang2020one}.  This family of approaches leverage Graph Neural Networks~\citep{gori2005new,scarselli2005graph} to condition the policy on the morphological representation of the agent, and have demonstrated the ability to generalize to unseen morphologies~\citep{huang2020one,wang2018nervenet,sanchez2018graph}.  Transformers~\citep{vaswani2017attention}, have also been adopted to further improve the task generalization in deep reinforcement learning~\citep{kurin2020my}, mainly by exploiting the attention mechanism for structure modelling~\citep{tenney2019you,vig2019analyzing}.

\section{Experiments}

\subsection{The imujoco dataset}

A few benchmarks have been proposed to address meta-learning and offline learning in RL, such as Meta-World~\citep{yu2020meta}, Procgen~\citep{cobbe2020leveraging}, and D4RL~\citep{fu2020d4rl}. However, differently from the standard meta-learning setting, in imitation learning we need a large variety of offline trajectories, collected from policies trained on heterogeneous environments. Existing benchmarks are not suited for this case as they: do not provide pretrained policies and their associated trajectories (e.g. Meta-World and Procgen), lack in diversity (Meta-World and D4RL), or do not support continuous control problems (e.g. Procgen).
In order to satisfy these requirements, we created a variant of OpenAI-Gym MuJoCo that we called iMuJoCo (iMitation MuJoCo). The OpenAI-Gym MuJoCo suite \citep{todorov2012mujoco, brockman2016openai} is a set of 10 control environments that has been widely used by the RL  community. This suite is useful for training standard online agents, but is less useful for studying imitation learning, due to its lack in heterogeneity.
The iMuJoCo dataset builds on top of MuJoCo providing a heterogeneous benchmark for training and testing imitation learning methods and offline RL methods. Heterogeneity is achieved by producing a large number of variants of three base environments: Hopper, Halfcheetah, and Walker2d. For each variant a policy has been trained via SAC, then the policy has been used to generate 100 offline trajectories. The user can access the environment variant (via the OpenAI-Gym API and a XML configuration file), the offline trajectories (via a Python data loader), and the underlying SAC policy network (using the Stable Baselines API, \citealt{stablebaselines3}). Each environment variant falls into one of these four categories:

\begin{itemize}
    \item \emph{mass}: increase or decrease the mass of a limb by a percentage; e.g. if the mass is 2.5 and the percentage is $200\%$ then the new mass for that limb will be $7.5$.
    \item \emph{joint}: limit the mobility of a joint by a percentage range, e.g. if the joint range is $180^{\circ}$ and the percentage is $-50\%$ then the maximum range of motion becomes $90^{\circ}$.
    \item \emph{length}: increase or decrease the length of a limb by a percentage; e.g. if the length of a limb is 1.5 and the percentage is $150\%$ then the new length will be $3.75$.
    \item \emph{friction}: increase or decrease the friction by a percentage (only for body parts that are in contact with the floor); e.g. if the friction is 1.9 and the percentage is $-50\%$ then the new friction will be 0.95.
\end{itemize}

Note that each environment has unique dynamics and agent configurations, resulting in different numbers of variants. Specifically, we have 37 variants for Hopper, 53 for Halfcheetah, and 64 for Walker2d, making a total of 154 variants. For a comprehensive list of these variants and additional details about the dataset, please refer to Appendix~\ref{appendix_imujoco}.

\subsection{Setup}
The policies used by all methods are parameterized via a Multi-Layer Perceptron (MLP) with two hidden layers (256 units) with ReLU activations. During the imitation phase we iterate over the support data for 20 epochs, minimizing the $\ell_2$ loss between the prediction of the model and the true target. We use the Adam optimizer \citep{kingma2014adam} with weight decay of $10^{-5}$, and an initial learning rate of $10^{-3}$ which is divided by 2 at $50\%$ and $75\%$ of the total learning epochs. These hyper-parameters were empirically derived from a grid search. At evaluation time, each method is adapted on the support data of each environment variation. There are a total of 100 trajectories from the target policy for each variant. Based on the number of shots, a random subset of trajectories is selected (e.g. 1 trajectory in 1-shot, 10 trajectories in 10-shot, etc) and used as support set. The remaining trajectories are used to estimate the test loss. For all methods we repeat the experiments using three different seeds and provide the aggregated results.
We compare five different methods: Scratch, Fine-Tuning, Head Fine-Tuning, Meta-Learning, and Multi-Task. The Scratch baseline is a neural network initialized from scratch and directly fine-tuned on the support data (no pretraining). The Fine-Tuning baseline is the implementation of our method, as described in previous sections. The Head Fine-Tuning baseline is like the Fine-Tuning baseline, but only the parameters of the last linear layer are optimized, the remaining set of weights is kept equal to the pretrained values. The Meta-Learning baseline is trained using first-order MAML as described in \cite{finn2017one, finn2017model}. At evaluation time we adapt the model similarly to Fine-Tuning (20 epochs of gradient descent using Adam optimizer and initial learning rate of $10^{-3}$); empirically this guarantees better results and ensures a fair comparison across methods.
Finally, Multi-Task is a behavioral cloning baseline trained on the same data distribution used in Meta-Learning and adapted to a query task in the same way as Fine-Tuning.
To test the generalization of the meta-learner and the multi-task learner we use a k-fold validation procedure: for each environment we split the variants into macro-categories (mass, friction, joint, length), and meta-train on all of them but one, the excluded category is then used for meta-evaluation. All experiments were carried out on a single NVIDIA A100 GPU with 80GB of memory.

Note that, in this section we compare all methods in terms of adaptation loss (loss over the query set of unseen tasks), reporting the reward scores in Appendix~\ref{appendix_experiments_tabular}. While in standard RL the maximum reward is often used to compare models, in few-shot imitation learning the reward is not a good proxy to asses the capabilities of a method. This is due to the fact that the target policy may have a reward that is lower/higher than the imitation policy. For instance if the imitation policy already performs well in the environment variant (high reward), but the target policy perform poorly (low reward), then the imitation policy has to lower its performance in order to approximate the target effectively.

\begin{figure}[t]
\centering
\includegraphics[width=\textwidth]{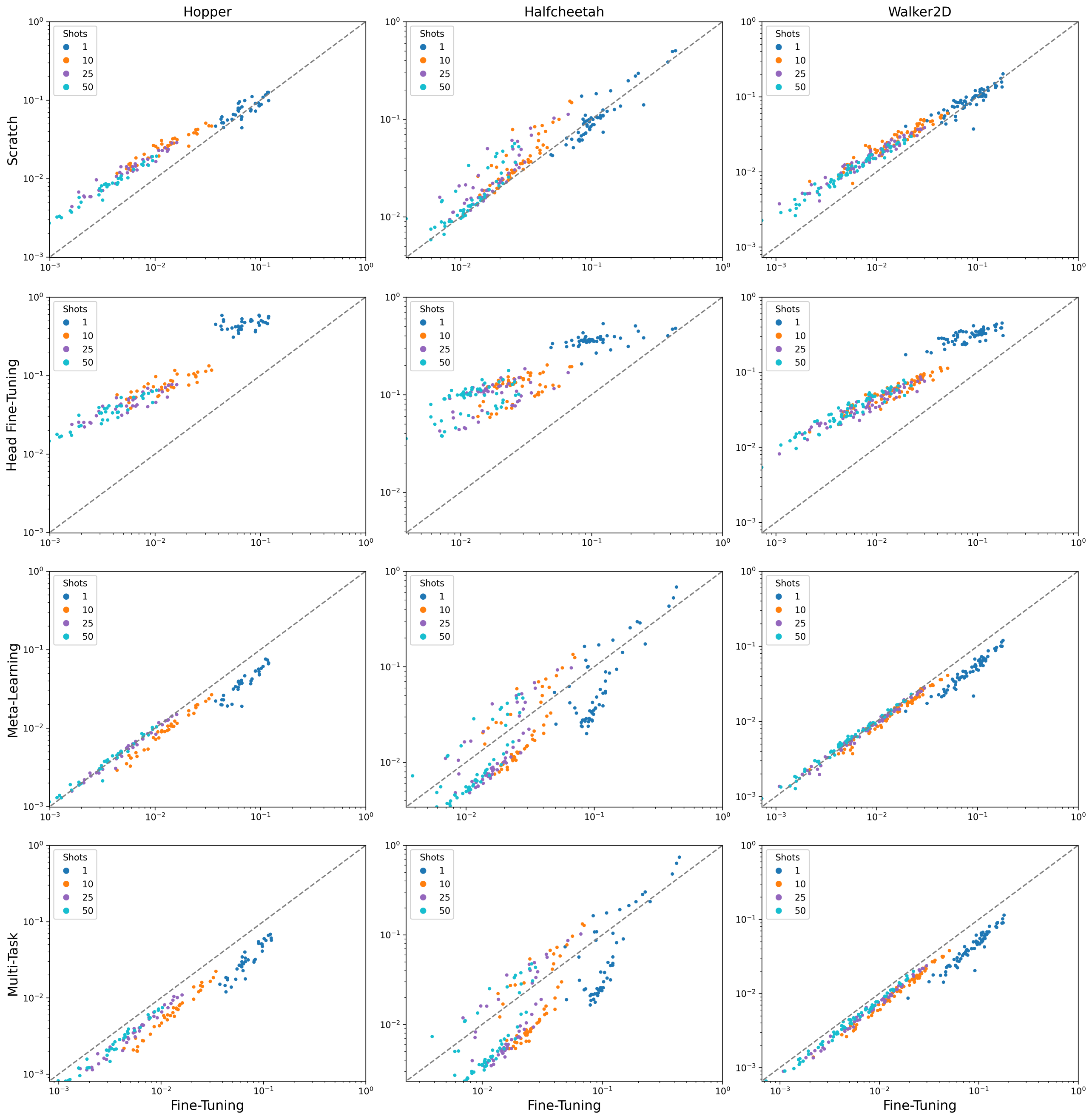}
\caption{Comparing the average query-set loss (log-scale) of Fine-Tuning against Scratch, Head Fine-Tuning, Meta-Learning, and Multi-Task over all shots (1, 10, 25, 50) and environments (Hopper, Halfcheetah, and Walker2d). Lower loss means a better approximation of the target policy. Each dot is the average loss over three runs for a particular environment variation. The dotted identity line represents equal performance (equal loss). Points above the line mean better performance for Fine-Tuning (lower loss for Fine-Tuning), whereas points below the line mean better performance for the method noted on the y-axis (lower loss for the other method).}
\label{fig_loss_scatter_all}
\end{figure}

\subsection{Comparing individual results}

In this section we examine the results reported in Figure~\ref{fig_loss_scatter_all} were we compare Fine-Tuning vs.~other methods in terms of adaptation capabilities on all environments variants and shots reporting the individual losses for all conditions. The idea is to get a bird-eye view of the results, before further breaking them down in the other sections. We report all the results in tabular format in Appendix~\ref{appendix_experiments_tabular}.

\textbf{Fine-Tuning vs.~Scratch} In order to asses whether the online pretraining stage is needed, we perform an ablation on the training protocol by removing the initial pretraining phase. We initialize a network from scratch for every task, and fine-tune it on the support set data. We refer to this variant as Scratch in the figures. The results are reported in the top row of Figure~\ref{fig_loss_scatter_all}, and show that Fine-Tuning performs better than Scratch in a large majority of conditions clearly showing the advantage of the pretraining stage. However, in the one-shot condition the two methods perform similarly, showing that a severe data scarcity may reduce the effectiveness of online pretraining at lower-shots.

\textbf{Fine-Tuning vs.~Head Fine-Tuning} The previous set of experiments confirms that pretraining is necessary to achieve good adaptation performance. However, fine-tuning a subset of the parameters may still be effective, as showed in the few-shot image classification setting \citep{bauer2017discriminative, tian2020rethinking, bronskill2021memory}. Here we verify this hypothesis by fine-tuning just the last linear layer of the pretrained backbone, calling this method Head Fine-Tuning. The results are reported in the second row of Figure~\ref{fig_loss_scatter_all}. Overall, fine-tuning the linear head significantly degrades the performance in all conditions. Interestingly, if we compare first and second row of Figure~\ref{fig_loss_scatter_all} we see that Head Fine-Tuning is less effective than Scratch. We hypothesize that this may be due to the fact that the neural networks used in RL are shallow, which reduce the reusability of the futures learned in early layers, this makes it necessary to adapt the whole set of parameters for obtaining the best performance.

\textbf{Fine-Tuning vs.~Meta-Learning} We compare Fine-Tuning and Meta-Learning, reporting the results in the third row of Figure~\ref{fig_loss_scatter_all}. The overall data show that the two methods perform very similarly on Hopper and Walker2d, while having complementary strengths and weaknesses on Halfcheetah. In the low-shot regime, in particular in 1-shot, we observe an advantage for Meta-Learning but this advantage disappears as we move towards higher shots. At 10-shot the methods perform similarly on Walker2D and Hopper, while on Halfcheetah results are mixed, with one method outperforming the other based on the variant considered. On high shots (25 and 50-shot) we observe that the two methods convergence towards similar performance, with a significant decrease in loss, meaning that both methods are able to effectively imitate the target policy.
These results suggest that, while Meta-Learning is more effective in the 1-shot case, the differences with Fine-Tuning tend to reduce substantially in the medium and high shot regimes. We conclude that at higher shots it may be more practical to use Fine-Tuning over Meta-Learning, bypassing expensive meta-training routines.

\textbf{Fine-Tuning vs.~Multi-Task} Finally, we compare Fine-Tuning and Multi-Task, reporting results in the last row of Figure~\ref{fig_loss_scatter_all}.
Overall Multi-Task is better than Fine-Tuning in most conditions but the gap in performance is smaller in Walker2d and (on average) in Halfcheetah. The general trend observed with Multi-Task resemble the one observed with Meta-Lerning, which may be due to the use of the same k-fold pretraining schedule. Notably, there is a performance advantage of Multi-Task over Meta-Learning, even though both methods have access to the same type of data. This suggests that Multi-Task may offer a more convenient approach in terms of training ease and performance improvement compared to Meta-Learning.
The performance advantage of Multi-Task over Fine-Tuning is likely due to the fact that Multi-Task has access to a larger variety of environments during the pretraining phase, while Fine-Tuning has only access to a single standard-environment. However, it is worth noting that assuming access to a distribution over tasks and policies may not be realistic in many practical applications, as this distribution may not be available. 

\begin{figure}[t]
\centering
\includegraphics[width=\textwidth]{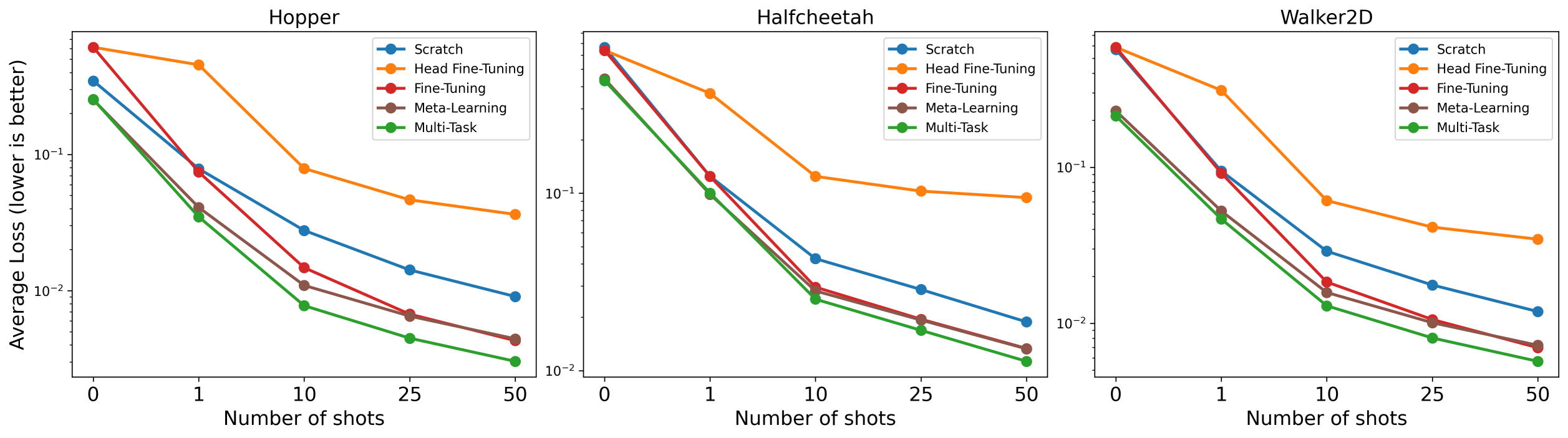}
\caption{Comparing the average query-set loss per shots (log-scale) for all methods and environments. Lower loss is better as it implies a better approximation of the target policy. As a reference, we include the 0-shot condition, which shows the loss before adaptation. The general trend show that for all methods the loss decreases as the number of shots increases. The difference between Fine-Tuning (red line) and Meta-Learning (brown line) becomes marginal in the medium (10-shot) and high-shot regime (25 and 50-shot). Fine-Tuning outperforms Scratch (blue line) in all conditions, showing the importance of the pretraining stage.}
\label{fig_loss_vs_shots_all}
\end{figure}

\subsection{Aggregated comparison}

In this section we analyze the aggregated loss per shot over all variants and for all methods. Results are reported in Figure~\ref{fig_loss_vs_shots_all}.

\textbf{Analysis of the 0-shot setting} The 0-shot condition is included as a baseline, showing how all methods perform \emph{before adaptation}. The performance of Head Fine-Tuning is identical to Fine-Tuning in this condition, as both use the same pretrained backbone which are not adapted at 0-shot. Fine-Tuning performs similarly to or worse than Scratch, with an average loss of $0.61$ vs.~$0.35$ in favor of Scratch for the Hopper environment, and similar scores on Halfcheetah ($0.64$ vs.~$0.66$) and Walker2d ($0.59$ vs.~$0.57$). These results confirm that the online pretraining stage does not provide an advantage for the fine-tuner. The situation is different for Meta-Learning, that does better at 0-shot in all three environments. The gap against Fine-Tuning is substantial in this condition with average losses of $0.35$ vs.~$0.25$ in Hopper, $0.64$ vs.~$0.44$ in Halfcheetah, and $0.57$ vs.~$0.22$ in Walker2d. Similar results are also observed when comparing Fine-Tuning and Multi-Task. We hypothesize that this is due to the fact that Meta-Learning and Multi-Task have been exposed to a large number of task variants during training while the other methods have seen only one or none. Therefore, the set of weights obtained by those methods may be more general-purpose at 0-shot. This implicit advantage of Meta-Learning and Multi-Task at 0-shot could also explain why they do better at 1-shot.

\textbf{Analysis of the low-shot setting} Here we focus on the low-shot conditions: 1-shot and 10-shot. The highest loss for all methods (worst imitation capabilities) is observed in the 1-shot condition (1 trajectory in the support set). This is expected, as the amount of data available in 1-shot is limited. In 1-shot we observe the largest gap between Fine-Tuning and Meta-Learning, with an average loss of $0.07$ vs.~$0.04$ on Hopper, $0.12$ vs.~$0.01$ on Halfcheetah, and $0.09$ vs.~$0.05$ on Walker2d. Multi-Task performs similarly to Meta-Learning or marginally better in this condition. Overall, Meta-learning and Multi-Task are more effective than Fine-Tuning in this scenario, but all methods suffer from poor approximation of the target policy due to the scarcity of data. In the 10-shot condition, we observe an overall decrease in loss (better approximation of the target policy) for all methods, showing the benefits of having more support data. The gap between Fine-Tuning and Meta-Learning decreases, with average losses of $0.014$ vs.~$0.011$ on Hopper, $0.030$ vs.~$0.028$ on Halfcheetah, and $0.018$ vs.~$0.016$ on Walker2d. At 10-shot Fine-Tuning outperforms both Scratch and Head Fine-Tuning by a substantial margin. The gap between Fine-Tuning and Multi-Task remains approximately constant on Hopper but it decreases in Halfcheetah and Walker2d.

\textbf{Analysis of the high-shot setting} Here we analyze the high-shot conditions: 25-shot and 50-shot. Overall we see a better performance for all methods as the number of shots increases. At 25-shot and 50-shot there is no significant difference between Fine-Tuning and Meta-Learning, with average losses being equal up to the fourth decimal place. Both methods outperform the others by a substantial margin. Comparing Fine-Tuning with Scratch we see that the former does better in all conditions; e.g. in the 50-shot condition, we observe a loss of $0.004$ vs.~$0.009$ on Hopper, $0.013$ vs.~$0.019$ on Halfcheetah, and $0.007$ vs.~$0.012$ on Walker2d. An important observation is that the gap between Scratch and Meta-Learning remains roughly constant as the number of shots increases, while the gap between Fine-Tuning an Meta-Learning gets smaller and eventually becomes marginal. This confirms the importance of online pretraining. The overall results for the Head Fine-Tuning method show that this approach has poor adaptation capabilities in all setting and environments. Finally, we observe that Multi-Task is the best performing method at higher shots. The gap with respect to Meta-Learning and Fine-Tuning increases pointing to the possibility that there is a qualitative difference in the optimum reached by this method.

\begin{figure}[t]
\centering
\includegraphics[width=\textwidth]{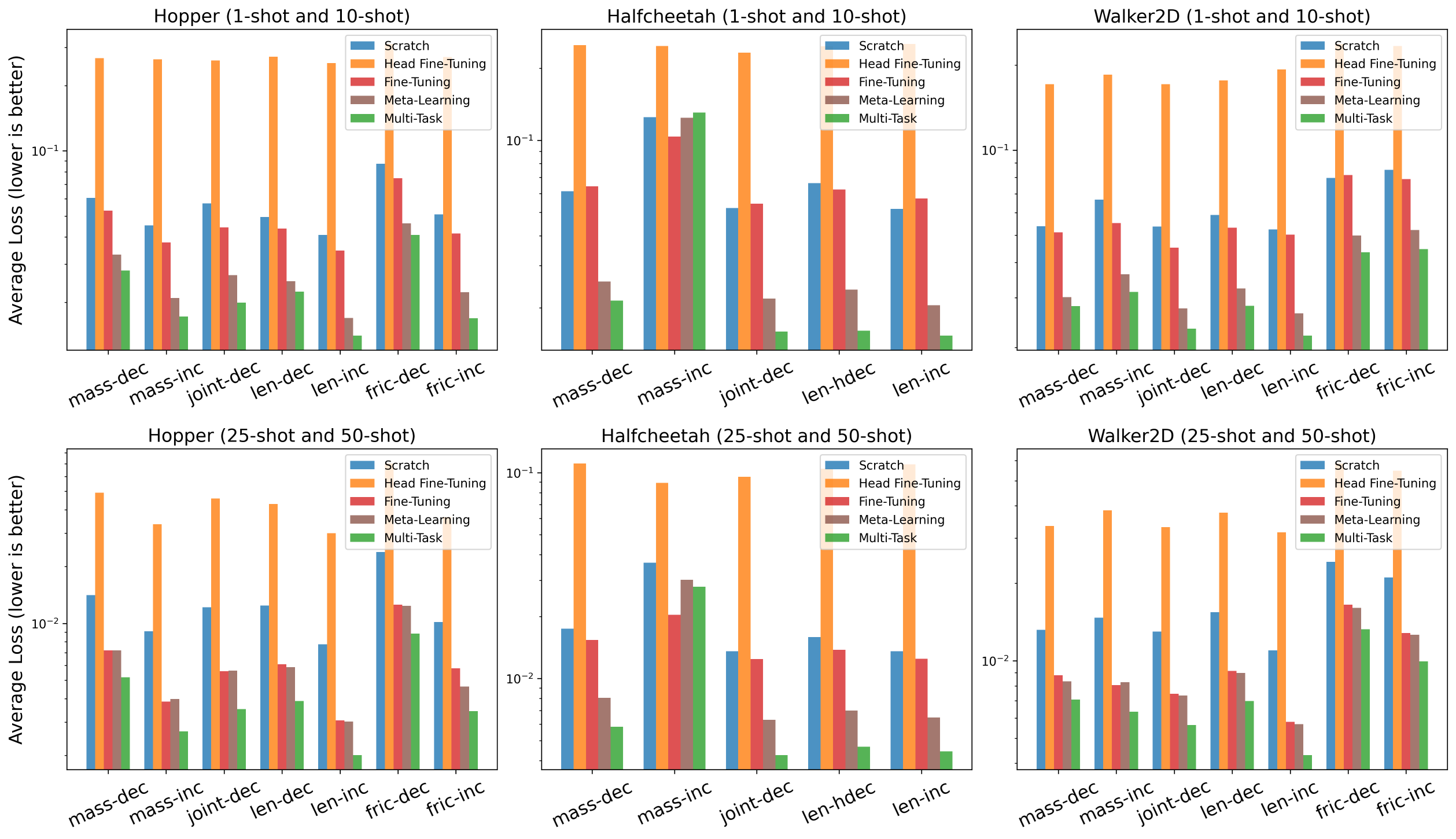}
\caption{Comparing the average query-set loss (log-scale) on low-shot (top row) and high-shots (bottom row) for all methods over subset of environment variants that include: decreasing/increasing limb mass (mass-dec, mass-inc), decreasing joint mobility (joint-dec), increasing/decreasing length of a limb (len-dec, len-inc), and increasing/decreasing the friction with the floor (fric-dec, fric-inc). Lower loss is better as it implies a better approximation of the target policy.}
\label{fig_env_variants_all_shots}
\end{figure}

\subsection{Comparison over environment variants}

In this section we breakdown the results by analyzing the performance over specific subsets of the iMuJoCo environment variants. In particular, we look at how methods perform with respect to modifications of the mass, joint mobility, length, and friction. Figure~\ref{fig_env_variants_all_shots} provides a summary of the results.

Overall we observe that some variants are more challenging than others but this is environment-related. For instance, on Hopper decreasing the friction seems to be the most difficult variant, as showed by the higher loss for all methods in both low and high-shot. In Halfcheetah increasing the mass is the most difficult variant, while decreasing the range of motion of the joints is the less difficult. In Walker2d, modifying the friction is the most problematic variant, whereas increasing the length of the joint produces the less damage.

Meta-Learning outperforms Fine-Tuning in most of the low-shot conditions, but the difference is marginal in high-shot conditions on Hopper and Walker2d, while being more pronounced on Halfcheetah. We do not observe any sensible trend related to environment variants (e.g. differences in performance across environments correlated to the same set of modifications). The only notable difference seems to be the lower loss for Fine-Tuning on variants that increases the mass of the limbs. This is particularly evident for the Halfcheetah environment, where Fine-Tuning outperforms Meta-Learning and Multi-Task by a substantial margin.

Finally, Fine-Tuning performs similarly to Scratch in the low-shot case, but the situation is reversed in the high-shot condition. This shows once again that the pretraining stage becomes more important at high-shot. 

\section{Conclusions}
In this paper we have investigated the few-shot imitation learning setting, proposing a simple yet effective fine-tuning baseline. We showed that by pretraining online via RL and fine-tuning offline on a set of support trajectories, it is possible to obtain results that are in line with meta-learning methods (in the high-shot regime) without the burden of meta-training on a distribution over tasks. In addition, we have released a dataset called iMuJoCo (iMitation MuJoCo) which includes 154 environment
variants obtained from the popular OpenAI-Gym MuJoCo suite with associated pretrained target policies and
trajectories. We believe that iMuJoCo can be a useful tool for investigating imitation learning and offline RL.

\textbf{Comparison with few-shot image classification} In the low-shot regime (1 and 10-shot), we have observed that meta-learning is a performant method able to provide solid results. However, the Multi-Task baseline is the best performing method in this regime, which is in line with what has been found in few-shot image classification \citep{chen2019closer, kolesnikov2020big, shysheya2023fit}. Therefore, when it is possible to generate data from many tasks (e.g.~when domain randomization is available) we suggest to rely on meta-learning or multi-task learning instead of fine-tuning as those methods will likely lead to better performance. In the high-shot regime (25 and 50-shot) we did not see any significant difference between meta-learning and fine-tuning. The Multi-Task baseline is the best performing method overall in this condition but the gap with respect to Fine-Tuning and Meta-Learning is sometimes marginal (e.g. Halfcheetah and Walker2d). We conclude that fine-tuning is a viable alternative to meta-learning and multi-task learning in this setting, having the additional advantage of not requiring an expensive meta-training protocol and a distribution over tasks or policies, which may be more practical in many real-world applications.
Another conclusion we have drawn from our experiments is that fine-tuning just the linear head does not work very well in few-shot imitation, this contrasts with results from the few-shot image classification literature \citep{bauer2017discriminative, tian2020rethinking}.
Two factors may explain these results: differences in the neural network backbone and differences in the available data. Regarding the first factor, there is a substantial difference between the type backbones used in the two settings. Exploiting large convolutional networks in image classification makes it easier to reuse previously learned features, whereas the shallow networks used in RL may need more substantial adjustments in parameter space to be effective. Regarding the second factor, the amount of available data can also play a crucial role, since in few-shot image classification there is plenty of data that can be used to pretrain the model, making it easier to generalize, while few-shot imitation typically has limited data, making generalization more challenging.

\textbf{Future work} In future work we would like to investigate other fine-tuning protocols, and verify if the type of RL method used in the online pretraining stage has an impact on the overall performance during the evaluation phase. In particular, the use of on-policy vs.~off-policy methods, parameters such as the entropy coefficient used in SAC, and the heterogeneity of the training environment, are factors that may play an important role in providing robust and general policies. Investigating these factors could shed more light on the differences between fine-tuning and meta-learning in the few-shot imitation setting.

\subsubsection*{Acknowledgments}
Funding in direct support of this work: 
Massimiliano Patacchiola and Richard E. Turner are supported by an EPSRC Prosperity Partnership EP/T005386/1 between the EPSRC, Microsoft Research, and the University of Cambridge.

\bibliography{collas2023_conference}
\bibliographystyle{collas2023_conference}

\clearpage
\appendix

\section{Additional details on the imujoco dataset} \label{appendix_imujoco}

The iMuJoCo dataset is a dataset of offline trajectories and pretrained policies that can be used to study imitation learning and offline RL. The iMuJoCo dataset is based on a subset of the 10 environments of the OpenAI-Gym MuJoCo suite \citep{todorov2012mujoco, brockman2016openai}. The subset of environments used in iMuJoCo consists of: Hopper, Halfcheetah, and Walker2d. We modify each based environment to create multiple variants. For each variant we train a target policy via SAC. We use the Stable Baselines 3 library \citep{stablebaselines3}, with default hyper-parameters for the SAC algorithm. We train for $10^6$ steps, using a clip range of $0.1$. From each SAC target policy we generate 100 trajectories and store them offline. Each trajectory includes tuples $(s_t, r_t, a_t)$ for all the steps taken (1000 steps per trajectory). Each variant is identified by a string that follows the convention \emph{type-quantity-part} where:

\begin{itemize}
    \item \textbf{type} refers to the variant type, it can be one of: \emph{massinc, massdec, jointdec, lengthinc, lengthdec, frictioninc, frictiondec}, as explained in the paper.
    \item \textbf{quantity} is a percentage value identifying how much the modification is affecting the part, e.g. \emph{massinc-100} means that the mass of that part has been increased by $100\%$.
    \item \textbf{part} refers to the body part affected by the modification; agents in each environment have a different number of body parts.
\end{itemize}

Note that each agent has a specific number of joints, and the dynamics of each environment are different. Therefore, the number of variants per each environment is different. We have 37 variants for Hopper, 53 for Halfcheetah, and 64 for Walker2d, for a total of 154.
Below we report the ID strings identifying each variant:

\textbf{Hopper} massdec-50-thighgeom, jointdec-50-thighjoint, massdec-50-leggeom, frictioninc-25-footgeom, massinc-200-torsogeom, jointdec-25-legjoint, massinc-100-leggeom, massinc-100-footgeom, jointdec-50-footjoint, massinc-100-torsogeom, lengthinc-150-footgeom, jointdec-50-footjoint, massinc-200-thighgeom, massinc-300-footgeom, frictiondec-25-footgeom, massinc-100-thighgeom, jointdec-25-footjoint, massinc-300-thighgeom, massdec-25-torsogeom, lengthinc-100-torsogeom, jointdec-25-thighjoint, massinc-200-footgeom, massdec-25-footgeom, massinc-200-leggeom, massdec-50-footgeom, lengthdec-50-torsogeom, massdec-25-thighgeom, frictioninc-50-footgeom, massdec-25-leggeom, lengthdec-50-footgeom, lengthinc-100-footgeom, frictiondec-50-footgeom, jointdec-50-legjoint, massinc-300-leggeom, massdec-50-torsogeom, lengthinc-50-footgeom, massinc-300-torsogeom.

\textbf{Halfcheetah} massinc-300-bshin, massinc-200-ffoot, massinc-200-fthigh, massdec-50-head, jointdec-50-fshin, jointdec-25-bshin, jointdec-25-ffoot, massinc-100-fthigh, massdec-50-fthigh, massinc-300-fthigh, massdec-25-bthigh, massinc-300-fshin, jointdec-50-fthigh, massinc-300-head, jointdec-50-ffoot, lengthinc-100-ffoot, jointdec-25-fthigh, massinc-300-ffoot, massinc-200-head, massinc-300-bthigh, massinc-200-bthigh, massinc-300-bfoot, lengthinc-50-head, massinc-200-fshin, massdec-50-fshin, lengthinc-50-bfoot, jointdec-50-bthigh, massinc-100-fshin, massinc-100-head, lengthinc-50-ffoot, massdec-25-fthigh, massinc-200-bshin, massdec-50-bfoot, massinc-200-bfoot, massdec-50-ffoot, massdec-25-head, massdec-25-ffoot, massinc-100-bfoot, lengthdec-50-head, massinc-100-bthigh, massdec-25-bfoot, jointdec-50-bshin, jointdec-50-bfoot, massdec-50-bshin, jointdec-25-bfoot, massdec-50-bthigh, massinc-100-bshin, jointdec-25-fshin, lengthinc-100-bfoot, massdec-25-bshin, jointdec-25-bthigh, massdec-25-fshin, massinc-100-ffoot.

\textbf{Walker2d} massdec-25-footleftgeom, massinc-300-thighgeom, jointdec-25-thighleftjoint, massinc-200-thighleftgeom, massinc-200-leggeom, lengthinc-200-footgeom, jointdec-50-legleftjoint, lengthinc-200-torsogeom, massinc-200-torsogeom, massdec-25-torsogeom, jointdec-25-legjoint, massdec-25-leggeom, frictioninc-25-footleftgeom, massinc-100-thighgeom, lengthdec-50-torsogeom, massinc-200-thighgeom, massinc-100-thighleftgeom, massdec-50-legleftgeom, lengthinc-100-footgeom, frictioninc-50-footgeom, massdec-25-thighleftgeom, massdec-50-footgeom, lengthinc-200-footleftgeom, massdec-50-torsogeom, lengthinc-100-footleftgeom, jointdec-25-footjoint, massdec-50-thighleftgeom, massinc-300-leggeom, massdec-50-thighgeom, massinc-300-footgeom, lengthinc-100-torsogeom, massinc-200-footleftgeom, massinc-200-footgeom, massdec-50-footleftgeom, jointdec-50-footjoint, massdec-25-footgeom, jointdec-50-legjoint, frictiondec-25-footleftgeom, lengthdec-50-footleftgeom, lengthdec-50-footgeom, frictiondec-50-footgeom, massinc-100-torsogeom, massinc-100-footleftgeom, frictiondec-50-footleftgeom, jointdec-50-thighleftjoint, massdec-50-leggeom, massinc-200-legleftgeom, massinc-300-thighleftgeom, jointdec-25-legleftjoint, massinc-100-footgeom, massdec-25-thighgeom, massinc-100-leggeom, massdec-25-legleftgeom, massinc-100-legleftgeom, jointdec-50-footleftjoint, jointdec-50-thighjoint, massinc-300-footleftgeom, massinc-300-legleftgeom, massinc-300-torsogeom, frictiondec-25-footgeom, frictioninc-25-footgeom, frictioninc-50-footleftgeom, jointdec-25-thighjoint, jointdec-25-footleftjoint.

\section{Implementation Details} \label{appendix_implementation_details}

We used Pytorch for implementing all the experiments. The full code will be released with an open-source license.

\textbf{Fine-Tuning} We pretrain the Fine-Tuning baseline via SAC on the base environment (no variation applied). We use the Stable Baselines 3 library \citep{stablebaselines3}, with default hyper-parameters for the SAC algorithm. We train for $10^6$ steps, using a clip range of $0.1$. At evaluation time we take the pretrained model for that environment, and optimize the $\ell_2$ loss over the support points, obtaining a task-specific set of parameters. We use the same fine-tuning protocol described in the paper (20 epochs, Adam optimizer).

\textbf{Head Fine-Tuning} We use the exact same configuration used in Fine-Tuning, but during evaluation we only optimize the last linear layer, using same number of epochs and optimizer used for Fine-Tuning. We have tried different hyper-parameters (e.g. learning rate schedule) but the one used for Fine-Tuning lead to the best results.

\textbf{Scratch} This method initialize the weight of the policy randomly via Glorot initialization \citep{glorot2010understanding} for each evaluation task. The network is then fine-tuned using the protocol described in the paper (20 epochs, Adam optimizer). 

\textbf{Meta-Learning} We use the 1-step MAML implementation used for regression problems as described in \cite{finn2017model}. We optimize over 100 tasks per meta-batch, using an inner learning rate of $0.01$, and a meta learning rate of $0.001$. We use Adam as meta-optimizer, and SGD as inner optimizer. During the evaluation phase, we adjust the weights of the policy via multiple optimization steps. We empirically found that applying the same adaptation schedule used for the other methods (20 epochs, Adam optimizer) lead to the best performance. Therefore, we used this adaptation scheduler in all experiments.

\textbf{Multi-Task} We pretrain the neural network for 50 epochs on the same k-fold data split used in Meta-Learning. We used Adap optimizer with a learning rate of $0.01$ that is divided by half at 25 and 37 epochs. During the adaptation phase we use the same adaptation schedule used for the other methods (20 epochs, Adam optimizer).

\clearpage
\section{Experiments: results in tabular format} \label{appendix_experiments_tabular}

\begin{table}[h]
\caption{Fine-Tuning in Hopper. Comparing average reward/loss per shot.}
\begin{center}
\begin{adjustbox}{width=0.9\columnwidth,center}
\begin{tabular}{lccccccccccc}
 & \multicolumn{6}{c}{\textbf{Reward}} & \multicolumn{5}{c}{\textbf{Loss}}\\
 \cmidrule(lr){2-7} \cmidrule(lr){8-12}
 & \textbf{target} & \textbf{0-shot} & \textbf{1-shot} & \textbf{10-shot} & \textbf{25-shot} & \textbf{50-shot} & \textbf{0-shot} & \textbf{1-shot} & \textbf{10-shot} & \textbf{25-shot} & \textbf{50-shot}\\
\hline
massdec-25-thigh & 4533.1 & 4566.4 & 3369.0 & 4156.6 & 4384.6 & 4528.0 & 0.5721 & 0.1023 & 0.0086 & 0.0045 & 0.0034 \\
massdec-25-legleft & 4049.4 & 4661.4 & 3088.8 & 3358.1 & 3982.2 & 4118.8 & 0.7401 & 0.115 & 0.0236 & 0.0156 & 0.012 \\
massdec-25-thighleft & 5112.1 & 4554.8 & 4119.5 & 4186.3 & 4742.0 & 4847.3 & 0.4931 & 0.106 & 0.0363 & 0.0252 & 0.0189 \\
massdec-25-footleft & 4814.2 & 4610.9 & 3036.5 & 4400.5 & 4736.0 & 4788.7 & 0.737 & 0.0611 & 0.0057 & 0.0028 & 0.0017 \\
massdec-25-leg & 4820.6 & 4658.9 & 4082.1 & 4576.1 & 4796.3 & 4857.9 & 0.4416 & 0.0316 & 0.0042 & 0.0021 & 0.0016 \\
massdec-25-foot & 5863.6 & 4659.8 & 3342.1 & 4440.6 & 5498.1 & 5796.8 & 0.5954 & 0.0606 & 0.0112 & 0.006 & 0.0038 \\
\hline
\end{tabular}
\end{adjustbox}
\end{center}
\end{table}

\begin{table}[h]
\caption{Fine-Tuning in Halfcheetah. Comparing average reward/loss per shot.}
\begin{center}
\begin{adjustbox}{width=1.0\columnwidth,center}
\begin{tabular}{lccccccccccc}
 & \multicolumn{6}{c}{\textbf{Reward}} & \multicolumn{5}{c}{\textbf{Loss}}\\
 \cmidrule(lr){2-7} \cmidrule(lr){8-12}
 & \textbf{target} & \textbf{0-shot} & \textbf{1-shot} & \textbf{10-shot} & \textbf{25-shot} & \textbf{50-shot} & \textbf{0-shot} & \textbf{1-shot} & \textbf{10-shot} & \textbf{25-shot} & \textbf{50-shot}\\
\hline
massdec-25-bthigh & 2617.1 & 1427.0 & 254.2 & 1840.1 & 2030.5 & 2324.6 & 0.6084 & 0.0966 & 0.0236 & 0.016 & 0.0113 \\
massdec-25-ffoot & 2607.3 & 2017.2 & 403.5 & 1699.7 & 1409.4 & 2239.9 & 0.6129 & 0.1096 & 0.0301 & 0.0201 & 0.014 \\
massdec-25-fthigh & 2647.0 & 1909.0 & 599.9 & 1853.5 & 2305.9 & 2477.1 & 0.6503 & 0.117 & 0.0286 & 0.0176 & 0.0122 \\
massdec-25-bshin & 2533.6 & 1025.7 & 1141.5 & 1819.2 & 1720.4 & 2186.0 & 0.5981 & 0.0899 & 0.021 & 0.0136 & 0.0095 \\
massdec-25-bfoot & 2511.3 & 1079.7 & 448.9 & 1686.8 & 1872.8 & 2367.8 & 0.5869 & 0.0873 & 0.0211 & 0.0124 & 0.0085 \\
massdec-25-head & 2986.3 & 4440.4 & 872.9 & 1368.7 & 2220.7 & 2632.5 & 0.6812 & 0.1171 & 0.034 & 0.0233 & 0.0165 \\
\hline
\end{tabular}
\end{adjustbox}
\end{center}
\end{table}

\begin{table}[h]
\caption{Fine-Tuning in Walker2d. Comparing average reward/loss per shot.}
\begin{center}
\begin{adjustbox}{width=1.0\columnwidth,center}
\begin{tabular}{lccccccccccc}
 & \multicolumn{6}{c}{\textbf{Reward}} & \multicolumn{5}{c}{\textbf{Loss}}\\
 \cmidrule(lr){2-7} \cmidrule(lr){8-12}
 & \textbf{target} & \textbf{0-shot} & \textbf{1-shot} & \textbf{10-shot} & \textbf{25-shot} & \textbf{50-shot} & \textbf{0-shot} & \textbf{1-shot} & \textbf{10-shot} & \textbf{25-shot} & \textbf{50-shot}\\
\hline
massdec-25-torso & 4652.4 & 4488.7 & 2787.1 & 3914.4 & 4471.4 & 4632.3 & 0.5214 & 0.1209 & 0.0288 & 0.0167 & 0.0096 \\
massdec-25-footleft & 4814.2 & 4610.9 & 3036.5 & 4400.5 & 4736.0 & 4788.7 & 0.737 & 0.0611 & 0.0057 & 0.0028 & 0.0017 \\
massdec-25-thighleft & 5112.1 & 4554.8 & 4119.5 & 4186.3 & 4742.0 & 4847.3 & 0.4931 & 0.106 & 0.0363 & 0.0252 & 0.0189 \\
massdec-25-leg & 4820.6 & 4658.9 & 4082.1 & 4576.1 & 4796.3 & 4857.9 & 0.4416 & 0.0316 & 0.0042 & 0.0021 & 0.0016 \\
massdec-25-legleft & 4049.4 & 4661.4 & 3088.8 & 3358.1 & 3982.2 & 4118.8 & 0.7401 & 0.115 & 0.0236 & 0.0156 & 0.012 \\
massdec-25-foot & 5863.6 & 4659.8 & 3342.1 & 4440.6 & 5498.1 & 5796.8 & 0.5954 & 0.0606 & 0.0112 & 0.006 & 0.0038 \\
\hline
\end{tabular}
\end{adjustbox}
\end{center}
\end{table}

\begin{table}[h]
\caption{Head Fine-Tuning in Hopper. Comparing average reward/loss per shot.}
\begin{center}
\begin{adjustbox}{width=1.0\columnwidth,center}
\begin{tabular}{lccccccccccc}
 & \multicolumn{6}{c}{\textbf{Reward}} & \multicolumn{5}{c}{\textbf{Loss}}\\
 \cmidrule(lr){2-7} \cmidrule(lr){8-12}
 & \textbf{target} & \textbf{0-shot} & \textbf{1-shot} & \textbf{10-shot} & \textbf{25-shot} & \textbf{50-shot} & \textbf{0-shot} & \textbf{1-shot} & \textbf{10-shot} & \textbf{25-shot} & \textbf{50-shot}\\
\hline
massdec-25-thigh & 3672.3 & 3345.5 & 3395.0 & 2605.5 & 2450.0 & 2187.7 & 0.6338 & 0.4992 & 0.097 & 0.052 & 0.0392 \\
massdec-25-leg & 3429.9 & 3473.5 & 3525.5 & 2673.2 & 2616.5 & 1655.0 & 0.7654 & 0.5671 & 0.1053 & 0.0689 & 0.0525 \\
massdec-25-foot & 3835.8 & 3600.7 & 3677.1 & 2869.3 & 2801.0 & 2292.6 & 0.565 & 0.4181 & 0.0653 & 0.0423 & 0.0339 \\
massdec-25-torso & 3337.4 & 3326.6 & 3345.3 & 3007.8 & 2750.3 & 2668.0 & 0.4914 & 0.3767 & 0.0817 & 0.0531 & 0.0453 \\
massdec-50-foot & 3892.0 & 3623.5 & 3696.1 & 3255.8 & 2661.0 & 2849.5 & 0.5965 & 0.4733 & 0.1172 & 0.075 & 0.0594 \\
massdec-50-thigh & 3330.1 & 3298.0 & 3305.0 & 2076.6 & 1830.8 & 1847.3 & 0.6131 & 0.4703 & 0.0991 & 0.0652 & 0.0549 \\
\hline
\end{tabular}
\end{adjustbox}
\end{center}
\end{table}

\begin{table}[h]
\caption{Head Fine-Tuning in Halfcheetah. Comparing average reward/loss per shot.}
\begin{center}
\begin{adjustbox}{width=1.0\columnwidth,center}
\begin{tabular}{lccccccccccc}
 & \multicolumn{6}{c}{\textbf{Reward}} & \multicolumn{5}{c}{\textbf{Loss}}\\
 \cmidrule(lr){2-7} \cmidrule(lr){8-12}
 & \textbf{target} & \textbf{0-shot} & \textbf{1-shot} & \textbf{10-shot} & \textbf{25-shot} & \textbf{50-shot} & \textbf{0-shot} & \textbf{1-shot} & \textbf{10-shot} & \textbf{25-shot} & \textbf{50-shot}\\
\hline
massdec-25-fshin & 2163.6 & 1844.3 & 929.0 & 42.3 & 515.0 & 1057.5 & 0.6773 & 0.3337 & 0.0586 & 0.0458 & 0.0416 \\
massdec-25-ffoot & 2607.3 & 2010.5 & 1178.9 & -121.6 & 318.7 & 350.4 & 0.6129 & 0.351 & 0.1463 & 0.1231 & 0.1144 \\
massdec-25-bthigh & 2617.1 & 1451.1 & 684.7 & -120.3 & 227.1 & 289.9 & 0.6084 & 0.3596 & 0.1311 & 0.1087 & 0.0999 \\
massdec-25-fthigh & 2647.0 & 1852.3 & 797.5 & -128.0 & 220.9 & 314.3 & 0.6503 & 0.3695 & 0.15 & 0.1261 & 0.1177 \\
massdec-25-bshin & 2533.6 & 1021.2 & 1268.7 & -67.5 & 153.0 & 118.4 & 0.5981 & 0.3575 & 0.1293 & 0.1067 & 0.0991 \\
massdec-25-head & 2986.3 & 4484.2 & 1734.3 & -69.8 & -86.7 & 168.2 & 0.6812 & 0.3946 & 0.1566 & 0.1293 & 0.1196 \\
\hline
\end{tabular}
\end{adjustbox}
\end{center}
\end{table}

\begin{table}[h]
\caption{Head Fine-Tuning in Walker2d. Comparing average reward/loss per shot.}
\begin{center}
\begin{adjustbox}{width=1.0\columnwidth,center}
\begin{tabular}{lccccccccccc}
 & \multicolumn{6}{c}{\textbf{Reward}} & \multicolumn{5}{c}{\textbf{Loss}}\\
 \cmidrule(lr){2-7} \cmidrule(lr){8-12}
 & \textbf{target} & \textbf{0-shot} & \textbf{1-shot} & \textbf{10-shot} & \textbf{25-shot} & \textbf{50-shot} & \textbf{0-shot} & \textbf{1-shot} & \textbf{10-shot} & \textbf{25-shot} & \textbf{50-shot}\\
\hline
massdec-25-thighleft & 5112.1 & 4556.5 & 3851.8 & 3158.8 & 3068.4 & 3083.3 & 0.4931 & 0.2564 & 0.0786 & 0.0552 & 0.049 \\
massdec-25-footleft & 4814.2 & 4612.0 & 4252.7 & 2683.3 & 3319.1 & 3146.1 & 0.737 & 0.335 & 0.0333 & 0.0209 & 0.0156 \\
massdec-25-torso & 4652.4 & 4490.0 & 4271.8 & 2861.7 & 2422.9 & 2380.5 & 0.5214 & 0.3267 & 0.0875 & 0.0568 & 0.0479 \\
massdec-25-legleft & 4049.4 & 4653.1 & 3165.4 & 2135.6 & 2636.9 & 2805.1 & 0.7401 & 0.3991 & 0.0682 & 0.0461 & 0.0403 \\
massdec-25-leg & 4820.6 & 4659.2 & 4129.1 & 3500.0 & 3084.6 & 2991.6 & 0.4416 & 0.1876 & 0.0201 & 0.0126 & 0.0096 \\
massdec-25-thigh & 4533.1 & 4566.9 & 4433.3 & 2518.6 & 3368.1 & 3373.9 & 0.5721 & 0.3137 & 0.046 & 0.0266 & 0.0223 \\
\hline
\end{tabular}
\end{adjustbox}
\end{center}
\end{table}

\begin{table}[h]
\caption{Scratch in Hopper. Comparing average reward/loss per shot.}
\begin{center}
\begin{adjustbox}{width=1.0\columnwidth,center}
\begin{tabular}{lccccccccccc}
 & \multicolumn{6}{c}{\textbf{Reward}} & \multicolumn{5}{c}{\textbf{Loss}}\\
 \cmidrule(lr){2-7} \cmidrule(lr){8-12}
 & \textbf{target} & \textbf{0-shot} & \textbf{1-shot} & \textbf{10-shot} & \textbf{25-shot} & \textbf{50-shot} & \textbf{0-shot} & \textbf{1-shot} & \textbf{10-shot} & \textbf{25-shot} & \textbf{50-shot}\\
\hline
massdec-25-foot & 3835.8 & 994.6 & 1683.9 & 1938.6 & 3059.5 & 3182.0 & 0.2974 & 0.0647 & 0.0243 & 0.0129 & 0.0085 \\
massdec-25-torso & 3337.4 & 1162.9 & 1875.4 & 1802.8 & 2512.6 & 2518.5 & 0.299 & 0.0725 & 0.0371 & 0.0223 & 0.017 \\
massdec-25-leg & 3429.9 & 839.0 & 1784.7 & 1916.5 & 2039.7 & 2431.8 & 0.3666 & 0.0987 & 0.0383 & 0.0176 & 0.0108 \\
massdec-25-thigh & 3672.3 & 491.0 & 1788.4 & 2567.3 & 2657.0 & 3398.2 & 0.3664 & 0.1108 & 0.0258 & 0.0107 & 0.0062 \\
massdec-50-thigh & 3330.1 & 855.9 & 2213.2 & 3350.3 & 2743.6 & 3098.6 & 0.2504 & 0.092 & 0.042 & 0.0228 & 0.0152 \\
massdec-50-foot & 3892.0 & 1041.4 & 1683.0 & 2162.9 & 3451.7 & 3381.4 & 0.3393 & 0.1166 & 0.0468 & 0.0235 & 0.0148 \\
\hline
\end{tabular}
\end{adjustbox}
\end{center}
\end{table}

\begin{table}[h]
\caption{Scratch in Halfcheetah. Comparing average reward/loss per shot.}
\begin{center}
\begin{adjustbox}{width=1.0\columnwidth,center}
\begin{tabular}{lccccccccccc}
 & \multicolumn{6}{c}{\textbf{Reward}} & \multicolumn{5}{c}{\textbf{Loss}}\\
 \cmidrule(lr){2-7} \cmidrule(lr){8-12}
 & \textbf{target} & \textbf{0-shot} & \textbf{1-shot} & \textbf{10-shot} & \textbf{25-shot} & \textbf{50-shot} & \textbf{0-shot} & \textbf{1-shot} & \textbf{10-shot} & \textbf{25-shot} & \textbf{50-shot}\\
\hline
massdec-25-fthigh & 2647.0 & -3.3 & -87.8 & 198.9 & 672.8 & 1901.8 & 0.6504 & 0.1091 & 0.035 & 0.022 & 0.0149 \\
massdec-25-bthigh & 2617.1 & -3.3 & 361.2 & 863.8 & 1265.6 & 2286.0 & 0.6731 & 0.08 & 0.0264 & 0.0166 & 0.0115 \\
massdec-25-bfoot & 2511.3 & -2.9 & 393.9 & 1246.0 & 1896.6 & 2295.1 & 0.6291 & 0.0701 & 0.0241 & 0.0135 & 0.0089 \\
massdec-25-bshin & 2533.6 & -3.5 & 819.9 & 1450.6 & 1573.5 & 2200.3 & 0.605 & 0.0715 & 0.0243 & 0.0157 & 0.0104 \\
massdec-25-head & 2986.3 & -2.4 & -126.7 & 1203.2 & 1599.2 & 1858.4 & 0.6798 & 0.1027 & 0.0422 & 0.028 & 0.019 \\
massdec-25-ffoot & 2607.3 & -2.3 & -28.2 & 166.6 & 1618.6 & 1984.5 & 0.7169 & 0.105 & 0.0373 & 0.0235 & 0.0157 \\
\hline
\end{tabular}
\end{adjustbox}
\end{center}
\end{table}

\begin{table}[h]
\caption{Scratch in Walker2d. Comparing average reward/loss per shot.}
\begin{center}
\begin{adjustbox}{width=1.0\columnwidth,center}
\begin{tabular}{lccccccccccc}
 & \multicolumn{6}{c}{\textbf{Reward}} & \multicolumn{5}{c}{\textbf{Loss}}\\
 \cmidrule(lr){2-7} \cmidrule(lr){8-12}
 & \textbf{target} & \textbf{0-shot} & \textbf{1-shot} & \textbf{10-shot} & \textbf{25-shot} & \textbf{50-shot} & \textbf{0-shot} & \textbf{1-shot} & \textbf{10-shot} & \textbf{25-shot} & \textbf{50-shot}\\
\hline
massdec-25-foot & 5863.6 & 175.6 & 308.4 & 1249.0 & 2802.8 & 4164.1 & 0.6355 & 0.0547 & 0.0164 & 0.0099 & 0.0064 \\
massdec-25-torso & 4652.4 & 171.1 & 1382.2 & 343.7 & 2502.3 & 3167.6 & 0.4405 & 0.1002 & 0.0379 & 0.024 & 0.0151 \\
massdec-25-legleft & 4049.4 & 339.1 & -106.1 & 1417.4 & 2173.5 & 3659.9 & 0.4278 & 0.0899 & 0.0302 & 0.0201 & 0.0154 \\
massdec-25-leg & 4820.6 & 411.0 & -29.0 & 383.4 & 2860.4 & 3210.7 & 0.5622 & 0.0479 & 0.0088 & 0.0051 & 0.0037 \\
massdec-25-thigh & 4533.1 & 87.8 & 1601.2 & 1424.3 & 2208.8 & 3764.6 & 0.6173 & 0.1033 & 0.0167 & 0.0089 & 0.0063 \\
massdec-25-thighleft & 5112.1 & 1029.8 & 1505.0 & 1547.9 & 2728.3 & 3480.2 & 0.612 & 0.1172 & 0.0428 & 0.0304 & 0.0239 \\
\hline
\end{tabular}
\end{adjustbox}
\end{center}
\end{table}

\begin{table}[h]
\caption{Meta-Learning in Hopper. Comparing average reward/loss per shot.}
\begin{center}
\begin{adjustbox}{width=1.0\columnwidth,center}
\begin{tabular}{lccccccccccc}
 & \multicolumn{6}{c}{\textbf{Reward}} & \multicolumn{5}{c}{\textbf{Loss}}\\
 \cmidrule(lr){2-7} \cmidrule(lr){8-12}
 & \textbf{target} & \textbf{0-shot} & \textbf{1-shot} & \textbf{10-shot} & \textbf{25-shot} & \textbf{50-shot} & \textbf{0-shot} & \textbf{1-shot} & \textbf{10-shot} & \textbf{25-shot} & \textbf{50-shot}\\
\hline
massdec-25-leg & 3429.9 & 2313.5 & 2306.2 & 2940.4 & 3197.5 & 3237.6 & 0.295 & 0.0676 & 0.0154 & 0.0085 & 0.0051 \\
massdec-25-torso & 3337.4 & 2535.1 & 2164.5 & 2962.3 & 3238.2 & 3304.9 & 0.1951 & 0.0534 & 0.0199 & 0.0125 & 0.0097 \\
massdec-25-thigh & 3672.3 & 2540.1 & 2630.6 & 3157.2 & 3373.8 & 3530.7 & 0.2727 & 0.0423 & 0.0072 & 0.0041 & 0.0029 \\
massdec-25-foot & 3835.8 & 2140.4 & 2675.5 & 3207.6 & 3574.2 & 3785.1 & 0.2089 & 0.0337 & 0.0093 & 0.0061 & 0.0042 \\
massdec-50-foot & 3892.0 & 1939.9 & 2452.0 & 3618.7 & 3705.9 & 3755.1 & 0.2996 & 0.0758 & 0.0266 & 0.0128 & 0.0082 \\
massdec-50-leg & 3487.1 & 2320.0 & 2813.4 & 3064.0 & 2999.5 & 3156.2 & 0.1889 & 0.0406 & 0.0166 & 0.0085 & 0.0046 \\
\hline
\end{tabular}
\end{adjustbox}
\end{center}
\end{table}

\begin{table}[h]
\caption{Meta-Learning in Halfcheetah. Comparing average reward/loss per shot.}
\begin{center}
\begin{adjustbox}{width=1.0\columnwidth,center}
\begin{tabular}{lccccccccccc}
 & \multicolumn{6}{c}{\textbf{Reward}} & \multicolumn{5}{c}{\textbf{Loss}}\\
 \cmidrule(lr){2-7} \cmidrule(lr){8-12}
 & \textbf{target} & \textbf{0-shot} & \textbf{1-shot} & \textbf{10-shot} & \textbf{25-shot} & \textbf{50-shot} & \textbf{0-shot} & \textbf{1-shot} & \textbf{10-shot} & \textbf{25-shot} & \textbf{50-shot}\\
\hline
massdec-25-ffoot & 2607.3 & 2091.4 & 1969.0 & 2299.4 & 2397.5 & 2450.3 & 0.1772 & 0.0401 & 0.0149 & 0.0103 & 0.0077 \\
massdec-25-bshin & 2533.6 & 2239.9 & 2292.2 & 2397.7 & 2446.6 & 2471.4 & 0.079 & 0.0239 & 0.0089 & 0.006 & 0.0047 \\
massdec-25-bthigh & 2617.1 & 2338.8 & 2430.1 & 2523.0 & 2563.6 & 2593.9 & 0.0996 & 0.0271 & 0.0109 & 0.0075 & 0.0057 \\
massdec-25-bfoot & 2511.3 & 2217.2 & 2266.7 & 2407.0 & 2449.3 & 2456.1 & 0.0683 & 0.02 & 0.0084 & 0.0057 & 0.0043 \\
massdec-25-fthigh & 2647.0 & 2297.7 & 2056.5 & 2323.2 & 2379.5 & 2436.5 & 0.1812 & 0.039 & 0.0138 & 0.0097 & 0.0074 \\
massdec-25-head & 2986.3 & 2430.8 & 1873.7 & 2393.2 & 2510.5 & 2651.2 & 0.242 & 0.0529 & 0.0212 & 0.0144 & 0.0107 \\
\hline
\end{tabular}
\end{adjustbox}
\end{center}
\end{table}

\begin{table}[h]
\caption{Meta-Learning in Walker2d. Comparing average reward/loss per shot.}
\begin{center}
\begin{adjustbox}{width=1.0\columnwidth,center}
\begin{tabular}{lccccccccccc}
 & \multicolumn{6}{c}{\textbf{Reward}} & \multicolumn{5}{c}{\textbf{Loss}}\\
 \cmidrule(lr){2-7} \cmidrule(lr){8-12}
 & \textbf{target} & \textbf{0-shot} & \textbf{1-shot} & \textbf{10-shot} & \textbf{25-shot} & \textbf{50-shot} & \textbf{0-shot} & \textbf{1-shot} & \textbf{10-shot} & \textbf{25-shot} & \textbf{50-shot}\\
\hline
massdec-25-leg & 4820.6 & 2867.9 & 1974.4 & 2831.6 & 4732.3 & 4861.4 & 0.1734 & 0.0173 & 0.0037 & 0.0022 & 0.0018 \\
massdec-25-thighleft & 5112.1 & 2870.7 & 1887.8 & 3636.5 & 4373.1 & 4567.1 & 0.1842 & 0.0671 & 0.03 & 0.0226 & 0.0185 \\
massdec-25-torso & 4652.4 & 2790.7 & 1792.2 & 3769.5 & 4073.7 & 4496.5 & 0.2022 & 0.0638 & 0.0225 & 0.0145 & 0.0094 \\
massdec-25-foot & 5863.6 & 2923.0 & 1836.8 & 3677.2 & 5049.9 & 5490.7 & 0.2058 & 0.0308 & 0.0083 & 0.0051 & 0.0036 \\
massdec-25-footleft & 4814.2 & 2869.7 & 1343.7 & 4123.9 & 4489.7 & 4714.5 & 0.3206 & 0.0267 & 0.0045 & 0.0026 & 0.0017 \\
massdec-25-legleft & 4049.4 & 2874.2 & 1962.7 & 3187.8 & 3884.8 & 4122.9 & 0.2983 & 0.0659 & 0.0198 & 0.0144 & 0.0115 \\
\hline
\end{tabular}
\end{adjustbox}
\end{center}
\end{table}

\begin{table}[h]
\caption{Multi-Task in Hopper. Comparing average reward/loss per shot.}
\begin{center}
\begin{adjustbox}{width=1.0\columnwidth,center}
\begin{tabular}{lccccccccccc}
 & \multicolumn{6}{c}{\textbf{Reward}} & \multicolumn{5}{c}{\textbf{Loss}}\\
 \cmidrule(lr){2-7} \cmidrule(lr){8-12}
 & \textbf{target} & \textbf{0-shot} & \textbf{1-shot} & \textbf{10-shot} & \textbf{25-shot} & \textbf{50-shot} & \textbf{0-shot} & \textbf{1-shot} & \textbf{10-shot} & \textbf{25-shot} & \textbf{50-shot}\\
\hline
massdec-25-torso & 3337.4 & 2967.8 & 2879.6 & 3241.3 & 3281.9 & 3074.5 & 0.2117 & 0.0517 & 0.016 & 0.0093 & 0.0071 \\
massdec-25-leg & 3429.9 & 3344.1 & 2743.1 & 3295.9 & 3270.3 & 3291.5 & 0.2608 & 0.057 & 0.012 & 0.0057 & 0.0034 \\
massdec-25-foot & 3835.8 & 3204.8 & 2700.5 & 3539.8 & 3665.0 & 3798.9 & 0.1791 & 0.0268 & 0.0064 & 0.0044 & 0.003 \\
massdec-25-thigh & 3672.3 & 3187.3 & 2846.4 & 3418.9 & 3359.9 & 3462.6 & 0.3035 & 0.0319 & 0.0047 & 0.0028 & 0.0021 \\
massdec-50-leg & 3487.1 & 3390.7 & 2757.8 & 2812.1 & 3225.4 & 3301.7 & 0.1792 & 0.0398 & 0.0139 & 0.0065 & 0.0033 \\
massdec-50-torso & 3362.1 & 2711.8 & 2962.0 & 3341.3 & 3364.0 & 3358.1 & 0.1899 & 0.0297 & 0.0043 & 0.0026 & 0.0018 \\
\hline
\end{tabular}
\end{adjustbox}
\end{center}
\end{table}

\begin{table}[h]
\caption{Multi-Task in Halfcheetah. Comparing average reward/loss per shot.}
\begin{center}
\begin{adjustbox}{width=1.0\columnwidth,center}
\begin{tabular}{lccccccccccc}
 & \multicolumn{6}{c}{\textbf{Reward}} & \multicolumn{5}{c}{\textbf{Loss}}\\
 \cmidrule(lr){2-7} \cmidrule(lr){8-12}
 & \textbf{target} & \textbf{0-shot} & \textbf{1-shot} & \textbf{10-shot} & \textbf{25-shot} & \textbf{50-shot} & \textbf{0-shot} & \textbf{1-shot} & \textbf{10-shot} & \textbf{25-shot} & \textbf{50-shot}\\
\hline
massdec-25-bshin.xml & 2533.6 & 2141.7 & 2180.9 & 2406.0 & 2448.2 & 2486.2 & 0.0786 & 0.022 & 0.007 & 0.0046 & 0.0036 \\
massdec-25-ffoot.xml & 2607.3 & 1926.3 & 1778.5 & 2370.2 & 2466.2 & 2477.4 & 0.1407 & 0.0329 & 0.011 & 0.0073 & 0.0054 \\
massdec-25-head.xml & 2986.3 & 2237.0 & 1963.4 & 2423.8 & 2666.6 & 2822.5 & 0.2249 & 0.0454 & 0.016 & 0.0104 & 0.0075 \\
massdec-25-bthigh.xml & 2617.1 & 2142.1 & 2342.8 & 2544.1 & 2598.1 & 2594.2 & 0.086 & 0.024 & 0.0082 & 0.0054 & 0.004 \\
massdec-25-fshin.xml & 2163.6 & 1887.7 & 1867.3 & 2118.4 & 2145.4 & 2155.5 & 0.0971 & 0.0189 & 0.0054 & 0.0034 & 0.0025 \\
massdec-25-fthigh.xml & 2647.0 & 2105.7 & 1945.0 & 2443.9 & 2532.2 & 2575.6 & 0.1561 & 0.0314 & 0.0104 & 0.0068 & 0.0052 \\
\hline
\end{tabular}
\end{adjustbox}
\end{center}
\end{table}

\begin{table}[h]
\caption{Multi-Task in Walker2d. Comparing average reward/loss per shot.}
\begin{center}
\begin{adjustbox}{width=1.0\columnwidth,center}
\begin{tabular}{lccccccccccc}
 & \multicolumn{6}{c}{\textbf{Reward}} & \multicolumn{5}{c}{\textbf{Loss}}\\
 \cmidrule(lr){2-7} \cmidrule(lr){8-12}
 & \textbf{target} & \textbf{0-shot} & \textbf{1-shot} & \textbf{10-shot} & \textbf{25-shot} & \textbf{50-shot} & \textbf{0-shot} & \textbf{1-shot} & \textbf{10-shot} & \textbf{25-shot} & \textbf{50-shot}\\
\hline
massdec-25-foot & 5863.6 & 2375.7 & 2604.5 & 3995.7 & 5663.6 & 5797.3 & 0.2178 & 0.0304 & 0.0073 & 0.0043 & 0.0029 \\
massdec-25-footleft & 4814.2 & 2473.8 & 2003.9 & 4300.5 & 4775.4 & 4799.9 & 0.3992 & 0.0222 & 0.0033 & 0.0018 & 0.0012 \\
massdec-25-leg & 4820.6 & 2579.2 & 2464.5 & 4149.1 & 4776.0 & 4850.4 & 0.1617 & 0.0144 & 0.0028 & 0.0015 & 0.0013 \\
massdec-25-thighleft & 5112.1 & 2775.7 & 1324.3 & 4168.4 & 4668.5 & 4808.9 & 0.1777 & 0.0654 & 0.0281 & 0.0209 & 0.0173 \\
massdec-25-legleft & 4049.4 & 2457.4 & 1289.4 & 3439.4 & 3978.5 & 4030.6 & 0.285 & 0.0597 & 0.0172 & 0.0127 & 0.0101 \\
massdec-25-thigh & 4533.1 & 2793.6 & 3101.6 & 4267.2 & 4481.7 & 4523.1 & 0.1605 & 0.0466 & 0.0056 & 0.0031 & 0.0026 \\
\hline
\end{tabular}
\end{adjustbox}
\end{center}
\end{table}

\end{document}